\documentclass{article}
\usepackage{graphicx} 
\usepackage{pgfplots} 
\usepackage{fullpage}
\usepackage{siunitx}
\usepackage{tablefootnote}
\usepackage[hidelinks]{hyperref}
\usepackage{algorithm}
\usepackage{algpseudocode}
\usepackage[T1]{fontenc}

\newtheorem{experiment}{Experiment}[section]

\newtheorem{definition}[experiment]{Definition}

\author{Charlotte Knierim, Luca Versari, Robert Obryk, Blaise Ag\"uera y Arcas, Rif A. Saurous }
\date{Google, Paradigms of Intelligence Team \\ \today}
\title{BFF: Simple explanations for complex phenomena}

\begin{document}
\maketitle

\begin{abstract}
The ``Computational Life'' paper (Agüera y Arcas et al., 2024) argues that paired interactions in a computational soup are an effective way to find self-replicators. In this work, aided by recent developments in self-replicator detection, we explore the alternate hypothesis that self-replicators can be found  at least as easily using simple mutation random walks in program space. We also explore the claim that capping the maximum ``depth'' and ``width'' of the ancestry tree stops self-replicators from emerging, showing instead that it merely stops self-replicators from taking over the soup.

\end{abstract}

\section{Introduction}

In the ``Computational Life'' paper (Agüera y Arcas et al., 2024), self-replicators were detected by observing that they had ``taken over'' a computational soup. We observed this transition by measuring the size of a compressed version of the soup, which rapidly becomes significantly smaller as a self-replicator takes over the soup. In the interim, we have developed (reasonably) reliable direct self-replicator detectors. These detectors allow us to disentangle the initial appearance of self-replicators, which occurs in system with random mutation alone as well as ones with program interaction alone, and their widespread diffusion through the soup, which requires program interaction.

The work presented here suggests that, in BFF, distributionally tuned random mutation is at least as powerful as pairwise interactions for finding self-replicators, and potentially for evolving programs with specific purposes.

\subsection{Overview}
The BFF system consists of a soup of random strings of length 64 (also called tapes). Those tapes are interpreted as programs in BFF, a dialect of the Brainfuck language~\cite{bf93}. The I/O in BFF works via two heads that also sit on the tape, meaning that programs can modify themselves while they run. 

The system runs by repeatedly choosing two programs uniformly at random, concatenating them and running the result, length 128, as if it were a single program.\footnote{We have an upper bound on the number of steps in case the resulting program has an infinite loop.} After execution, the tapes are taken apart and the two pieces are placed back into the soup in their new state.

Note that we have seen many different variants of BFF: different ways to map bytes to ops, slightly different operator sets, different soup sizes, interactions with all programs paired up simultaneously or one interaction at a time. This document aims to be generic in the BFF dialect; we repeated the experiments for multiple variants, with very similar results. The main variant we use, \texttt{bff\_selfmove}~\cite{bffselfmove}, only has one copy operation and automatically moves the read head on a copy.

\subsection{Main Results}
Our primary finding is that we can easily find BFF self-replicators by executing a simple ``random walk mutation'' process in program space faster than we can find them by evolving the soup via paired computation. We measure time according to the number of programs seen during the process, and observe faster self-replicator emergence.

Statistics on the number of self-replicators in other computational systems and parallels to biology have been explored in \cite{10.1098/rsta.2016.0350}.

More generally, these experiments support the view that program interaction in BFF (or similar systems involving the dyadic interaction of short fixed-length tapes) is not an unusually powerful search operator. We continue to believe, for substantial theoretical and empirical reasons, that biology makes powerful use of recombination (both standard crossover and horizontal gene transfer (HGT)). However, we do not currently consider program interaction in BFF to be an effective model of these processes. Finding good computational instantiations of crossover and HGT remains an open problem.

As a second finding we present an analysis of compositionality in the BFF system. Even if the system does not find self-replicators faster than a random process, if we could see that self-replicators emerge in a compositional way that would be an interesting connection to biological evolution. We define a \emph{merger} to be a consecutive copy of multiple bytes that have not been previously copied together. We conduct experiments that limit the \emph{depth} or \emph{width} of a merger. The depth of a merger captures the compositional complexity of the resulting string: a string of high depth implies that it is composed of strings that have been merged and copied previously. The width of a merger controls how much complexity a single step can add. We show that even if we block low-depth and low-width mergers,
the system still generates self-replicators, although blocking merges does stop them from taking over the soup.  

We start this document with a short section that introduces how our self-replicator detection works. Afterwards, we describe the experiments we conducted and the conclusions drawn from them.

\section{Self-replication detector}
\label{sec:detector}

To better identify whether there are self-replicators in the soup, we developed a self-replication detector.

Intuitively, we want to test that a given program produces children that are identical to itself independently of what the program it is with (as long as the program is the first one in the pair). However, an exact comparison is not what we want: some self-replicators copy their functionality and most of their bytes but have some non-executed bytes that are kept from the original program. Similarly, some self-replicators invert themselves and will produce an exact copy only after an even number of executions. 
The following is a pseudo-code version of our self-replication detector \cite{cubff}:

\algdef{SE}[REPEATN]{RepeatN}{End}[1]{\algorithmicrepeat\ #1 \textbf{times}}{\algorithmicend}

\begin{algorithm}
\caption{Self-replication detection}
\begin{algorithmic}
    \State Input: Program $P$
    \For{$i\in [0..9]$}
    \State Initialize tape $T_i$ with the first half $T_{i,1}$ and the second half $T_{i,2}$
     \State $T_{i,2} \gets P$
    \RepeatN{5}

        \State  $T_{i,1} \gets T_{i,2}$
        \State $T_{i,2} \gets Noise$
        \State Run tape $T_i$
    \End
    \EndFor
    \State Compare the results (see text)
    \end{algorithmic}
\end{algorithm}

\begin{figure}
    \centering
    \includegraphics[width=\linewidth]{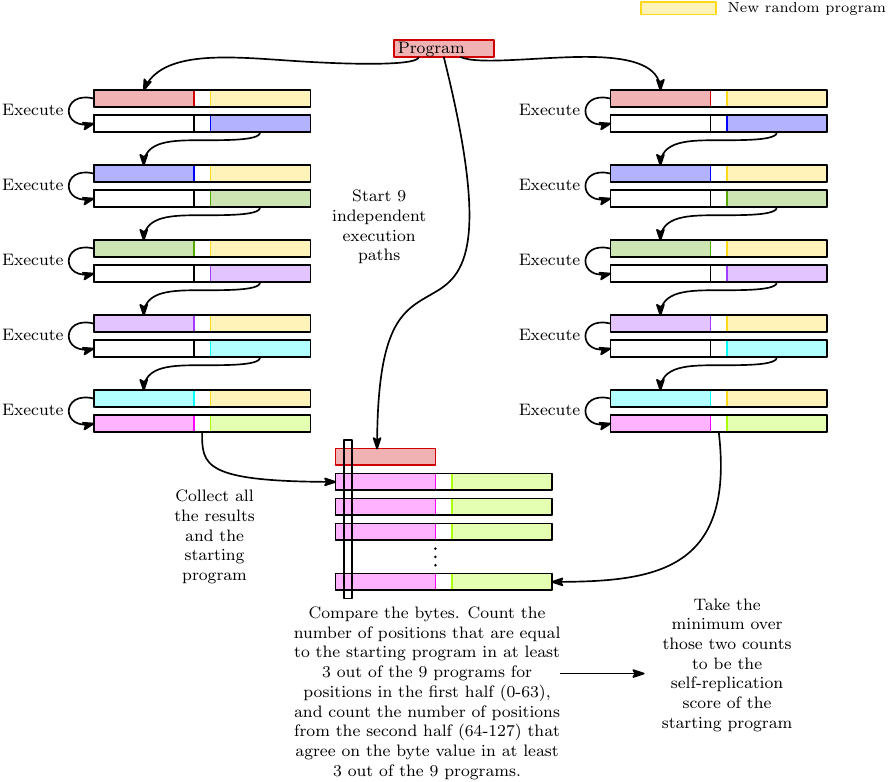}
    \caption{A schematic drawing of the self-replication detector}
    \label{fig:detector}
\end{figure}

For the result comparison, we collect the nine final tapes together and compare their halves separately. For the first halves, we compute the number of bytes that match the original program in at least three of the final tapes. For the second halves, we count the number of bytes that are identical in at least three of the resulting tapes. We do not compare the second halves to the first halves, as there are self-replicators that invert themselves in each iteration. We take the minimum score over the first and the second half and use that as the \emph{self-replication score}. Note that in most cases the comparison of the first halves will be the minimum and the comparison of the second halves can be seen as extra robustness against breaking when encountering a \texttt{0} during execution. A schematic drawing of the process is depicted in Figure~\ref{fig:detector}.

The basic intuition is that a self-replicator can replicate itself reliably and repeatably. The parameters are chosen somewhat arbitrarily: small enough to have a fast runtime, yet large enough to avoid false positives. We extend the chain of running the programs to a length of five to ensure it keeps being functional after executing itself, we have different branches to ensure the self-replicator is robust. We check for bytes in the exact same position, so our detector does not fire for ``semi-replicators'' that copy themselves into a different position on the destination tape. Ensuring that the length of this chain is odd is important, as this allows us to detect self-replicators that invert themselves with every execution.

The \emph{self-replication score} is an integer between $0$ and $64$.
Note that requiring the program to have a score of $64$ would require the program to exactly self-replicate itself, which often does not happen: not all bytes are important for functionality, so it is acceptable for a self-replicator to leave some bytes of the program untouched.

We consider a program to be a self-replicator if the self-replication score is at least $48$. This is a somewhat arbitrary choice, and results do not change significantly for thresholds of at least $\approx20$. 
Intuitively, if the program is classified as a self-replicator, then the resulting tapes agree on a fraction of the bytes that is much higher than sampling nine random tapes would explain. In this case, the program must have successfully written bytes in a way such that even after five iterations it is still functional, which is rare for anything that is not a self-replicator\footnote{One program that gets a very high score when testing for self replication is a program that consists of the alternating byte sequence $128,130$ with a minimal copy program (4 bytes long) in positions 3 to 6. This way, as the head positions are two apart, in each execution the program shifts itself by two positions. As most of the program is invariant to this shift, this causes 56 bytes to still be the same as in the original program but eventually the functional copy mechanism will wrap around to the front and be destroyed}.

\section{BFF experiments}
In this section, we describe experiments in the BFF setting. Most of the experiments replace the standard BFF interactions with a random process that alters the tapes.

We will compare BFF's execution model to a simpler model which mutates individual programs. Of course, a random process that treats programs as independent entities will never lead to the soup being taken over.  But if we compare the BFF system to the creation of life, then we are much more interested in the moment the first living being or self-replicator emerges. The development of our self-replicator detector makes this measurement possible.

To compare how long a system takes to self-replication is we consider two different metrics. In the first approach, we compare the number of \emph{programs tested for self-replication}. That is, a single interaction potentially changes both programs so we re-test both programs after every interaction, meaning that each interaction counts as two tested programs. In the second approach, we measure the \emph{number of bytes changed} in the system. This metric might seem like the more fair comparison if we think that sampling the program from scratch every time injects too much entropy into the system.
Both of these analyses take advantage of our self-replicator detector.

\subsection{Sampling random programs}
\label{sec:sampling}

As a baseline, we tested how long running the BFF system takes to see the first self-replicator. While previous work (see \cite{alakuijala2024computational}) has always looked at the point the soup \emph{gelates} (takes over the soup), we instead use our self-replication test. This also allows us to observe situations where a self-replicator appears but dies out before taking over a significant fraction of the soup.

All sampling experiments in this section were conducted by sampling $10^{9}$ random programs. When comparing a beta prior to a beta posterior in this setting, the variance is on the order of at most $10^{3}$ and thus negligible compared to the differences in means we observe.

\begin{experiment}
    The average time to first self-replicator in BFF is $\approx 2.5 \cdot 10^6$ interactions, which corresponds to $5\cdot10^6$ programs tested. This was measured by taking the average over 100 seeds.
    \label{ex:run}
\end{experiment}

One easy way to compare BFF to a random process is to replace every interaction with a process that creates two completely new random tapes. In this setting, we simply sample programs uniformly at random until a self-replicator is found. 
\begin{experiment}

When sampling random byte strings of length 64, we see  $2.9\cdot10^7$ programs on average before finding a self-replicator.
\label{ex:random}
\end{experiment}

Comparing Experiments~\ref{ex:run} and \ref{ex:random}, we see that pure random sampling is about 6 times slower than running BFF.

Experiment~\ref{ex:random} sampled programs from a uniform distribution over bytes. However, only a small fraction of bytes are BFF operators. We now consider settings where we sample from a distribution better suited to find self-replicators.\footnote{To make a biological analogy, we imagine a ``soup'' that is enriched with the precursors of life relative to a completely random soup.} For this, we first looked at the distribution of bytes written by the BFF system. 

\begin{figure}[t]
    \centering
\begin{tikzpicture}
\begin{axis}[
	enlargelimits=0.05,
	ybar,
    bar width=0.01cm,
    width=\textwidth,
    height=5cm,
    scaled y ticks=false,
    xtick distance=32,
    y tick label style={/pgf/number format/1000 sep=\,},
    ymode=log,
    log basis y={2}
]
\addplot 
	coordinates {(0, 64746) (1, 26688) (2, 13654) (3, 11642) (4, 14072) (5, 15646) (6, 14200) (7, 7913) (8, 7006) (9, 6727) (10, 6785) (11, 6735) (12, 6529) (13, 6519) (14, 6451) (15, 6811) (16, 6789) (17, 6682) (18, 7012) (19, 6875) (20, 6480) (21, 6599) (22, 6725) (23, 6927) (24, 6729) (25, 6791) (26, 6512) (27, 6858) (28, 6912) (29, 6823) (30, 6655) (31, 6667) (32, 6874) (33, 6553) (34, 6586) (35, 6795) (36, 6702) (37, 6283) (38, 6559) (39, 6560) (40, 6823) (41, 6927) (42, 6533) (43, 6790) (44, 6648) (45, 6613) (46, 6518) (47, 6649) (48, 6419) (49, 6530) (50, 6616) (51, 6690) (52, 6724) (53, 6684) (54, 6628) (55, 6797) (56, 6584) (57, 6651) (58, 6652) (59, 6753) (60, 6593) (61, 6722) (62, 6446) (63, 6718) (64, 6637) (65, 6968) (66, 7054) (67, 6975) (68, 7104) (69, 6986) (70, 7172) (71, 6755) (72, 6940) (73, 7237) (74, 6977) (75, 7037) (76, 7222) (77, 7224) (78, 7009) (79, 6873) (80, 6855) (81, 6993) (82, 6730) (83, 6998) (84, 6955) (85, 7378) (86, 6470) (87, 6537) (88, 7029) (89, 6661) (90, 6835) (91, 6865) (92, 6994) (93, 6719) (94, 6649) (95, 6783) (96, 6744) (97, 6775) (98, 6915) (99, 7247) (100, 7002) (101, 7851) (102, 6765) (103, 6786) (104, 6944) (105, 6912) (106, 7225) (107, 6795) (108, 6863) (109, 6889) (110, 6759) (111, 6804) (112, 6922) (113, 6690) (114, 6800) (115, 7009) (116, 6960) (117, 7287) (118, 7252) (119, 6907) (120, 6799) (121, 6944) (122, 7125) (123, 7154) (124, 7502) (125, 7500) (126, 7431) (127, 8048) (128, 8832) (129, 10052) (130, 8241) (131, 7261) (132, 6966) (133, 6847) (134, 7434) (135, 6728) (136, 6992) (137, 6841) (138, 6797) (139, 6724) (140, 6850) (141, 6786) (142, 6804) (143, 6852) (144, 6898) (145, 6845) (146, 6853) (147, 6612) (148, 6688) (149, 6809) (150, 6730) (151, 6849) (152, 6865) (153, 6565) (154, 6669) (155, 6599) (156, 6510) (157, 6567) (158, 6604) (159, 6642) (160, 6441) (161, 6602) (162, 6666) (163, 6643) (164, 6413) (165, 6614) (166, 6584) (167, 6379) (168, 6733) (169, 6664) (170, 6938) (171, 6754) (172, 6542) (173, 6492) (174, 6411) (175, 7118) (176, 6584) (177, 6489) (178, 6643) (179, 6466) (180, 6559) (181, 6609) (182, 6604) (183, 6558) (184, 6625) (185, 6666) (186, 6728) (187, 6790) (188, 6755) (189, 6109) (190, 6460) (191, 6473) (192, 6635) (193, 6617) (194, 6873) (195, 6850) (196, 6657) (197, 6648) (198, 6570) (199, 6665) (200, 6625) (201, 6909) (202, 6999) (203, 6745) (204, 6617) (205, 7058) (206, 7209) (207, 7073) (208, 6748) (209, 6890) (210, 6858) (211, 6840) (212, 7040) (213, 6977) (214, 6733) (215, 6743) (216, 6723) (217, 6926) (218, 6917) (219, 7253) (220, 7186) (221, 7022) (222, 6795) (223, 7103) (224, 6765) (225, 6393) (226, 6827) (227, 6689) (228, 6852) (229, 6643) (230, 7204) (231, 7053) (232, 6783) (233, 7151) (234, 6931) (235, 6975) (236, 6722) (237, 6931) (238, 7090) (239, 7206) (240, 7037) (241, 7311) (242, 7031) (243, 6939) (244, 6844) (245, 7168) (246, 6988) (247, 6909) (248, 6772) (249, 7514) (250, 7260) (251, 7262) (252, 7640) (253, 8649) (254, 10728) (255, 18876)};
\end{axis}
\end{tikzpicture}
    \caption{Frequencies of lasting changes into a given character (logscale on y). Here, by ``lasting'' we mean that if a byte gets changed multiple times during the same interaction, we only consider the last of these changes.}

    \label{fig:dist}
\end{figure}
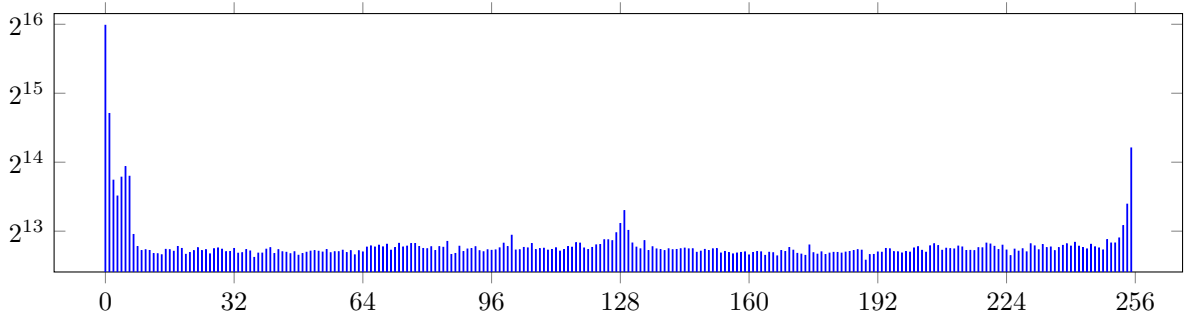

\begin{definition}
We define the distribution $\mathcal{BFF}$ to be the distribution where bytes are drawn according to the frequencies given in Figure~\ref{fig:dist}. This distribution was empirically measured by recording which bytes the BFF system writes shortly before a self-replicator takes over the soup.
\label{def:bffdist}    
\end{definition}

We now use this distribution to sample bytes when creating random bytes.

\begin{experiment}
When sampling random byte strings of length 64, where every byte is sampled from $\mathcal{BFF}$, we see $1.7\cdot10^6$ programs on average before finding a self-replicator.
\end{experiment}
This is roughly a factor of 3 faster in terms of programs tested than running the BFF system.

If the distribution discovered by BFF were the best distribution one could find, then we could argue that the value in running the BFF system lies in discovering this distribution and it might be hard to find a similarly good distribution manually or with a randomized strategy.
The next experiment shows that this is not the case.
\begin{definition}
Define the distribution $\mathcal{CUST}$ as follows: for each byte $b$, we flip a fair coin to decide whether $b$ is an operator or a no-op, then choose $b$ uniformly at random from the chosen class.
\label{def:custom}
\end{definition}
\begin{experiment}
When sampling random byte strings of length 64, where every byte is sampled from $\mathcal{CUST}$, we see $4.5\cdot10^5$ programs on average before we find a self-replicator.

\end{experiment}
This is a factor of 10 times more efficient at finding self-replicators than running BFF.

We introduce a slight optimization of this distribution, where we add the byte with value $64$ to the operator bucket. The rationale behind this is to make it easier to start with head positions that are exactly $64$ positions apart. Head positions that are exactly 64 apart imply that the two heads start in the same location in the first and second program respectively. This makes producing an exact copy of the first program much easier as there is no need to align the heads first.

\begin{definition}
    Define the distribution $\mathcal{CUST}64$ as follows: for each byte $b$, we flip a fair coin to decide whether $b$ is in the class of `operator or 64' or a no-op, then choose $b$ uniformly at random from the chosen class.
\label{def:custom64}
\end{definition}

\begin{experiment}
When sampling random byte strings of length 64, where every byte is sampled from $\mathcal{CUST}64$, we see $9.4\cdot10^4$ programs on average before we find a self-replicator.

\end{experiment}
This is roughly a factor of 25 times more efficient at finding self-replicators than running BFF.

One could argue that the experiments from this section change the system more than running BFF does and this explains the difference in speed. In the next section, we change our method from sampling whole programs to a random process that mutates every byte with a given probability $p$.

\subsection{Analyzing the number of changed bytes}
\label{sec:bytes}
In our next experiment, we measured the number of bytes changed in the BFF system in every interaction. We chose the mutation probability $p$ by measuring the number of changed bytes per interaction with varying rates that are similar to the rate of change we observed in the BFF system. 

The probability $p$ can be seen as a parameter chosen by the BFF system, similar to the distribution of characters we write. In \texttt{bff\_selfmove}, the fraction of bytes that change per interaction grows from $1/100$ to $\approx 1/50$ and then sharply increases as the soup gelates\footnote{For other BFF dialects, this value tends to stay constant at values in the range $1/200\to 1/100$ before increasing due to gelation. We have not looked into why different systems behave differently in this regard. However, we do not consider this crucial for this write-up.}. 

The experiments in this section look at the following random process:
 \begin{enumerate}
     \item Start with a random soup.
     \item\label{step:mmutate} Mutate every byte in the soup with probability $p$, according to a given distribution.
     \item\label{step:check} Check all the programs for self-replicators.
     \item Repeat Steps \ref{step:mmutate} and \ref{step:check} until we find a self-replicator.
 \end{enumerate}

 We ran this for different distributions and mutation probabilities. The results are summarized in Table~\ref{tab:replicator_time}. Note that the setup with $p=1$ is identical to the experiments in Section~\ref{sec:sampling}. All results in this table were obtained by testing $10^{10}$ programs each. For almost all of the results the variance is at least an order of magnitude smaller than the expected time to self-replication. For the uniform distribution with mutation probability of $1/100$ and $1/200$ the values are only about one variance apart and can thus not be ordered with certainty.
{\renewcommand{\arraystretch}{1.2}
\begin{table}[h!]
    \centering
    \caption{Expected time to find 1 replicator (Total Programs Tested / Replicators Found).}
    \label{tab:replicator_time}
    \begin{tabular}{|c|c|c|c|c|}
        \hline
        \textbf{Mutation Probability} & \textbf{Uniform} & \textbf{$\mathcal{BFF}$} & \textbf{$\mathcal{CUST}$} & \textbf{$\mathcal{CUST}64$} \\
        \hline
        $\frac{1}{200}$ & $9.5 \cdot 10^{8}$ & $3.5 \cdot 10^{7}$ & $7.9 \cdot 10^{6}$ & $1.9 \cdot 10^{6}$ \\
        $\frac{1}{100}$ & $4.4 \cdot 10^{8}$ & $1.9 \cdot 10^{7}$ & $4.1 \cdot 10^{6}$ & $9.9 \cdot 10^{5}$ \\
        $\frac{1}{50}$ & $2.0 \cdot 10^{8}$ & $1.0 \cdot 10^{7}$ & $2.1 \cdot 10^{6}$ & $5.2 \cdot 10^{5}$ \\
        $1$ & $2.9 \cdot 10^{7}$ & $1.7 \cdot 10^{6}$ & $4.5 \cdot 10^{5}$ & $9.4 \cdot 10^{4}$ \\
        \hline
    \end{tabular}
\end{table}
}

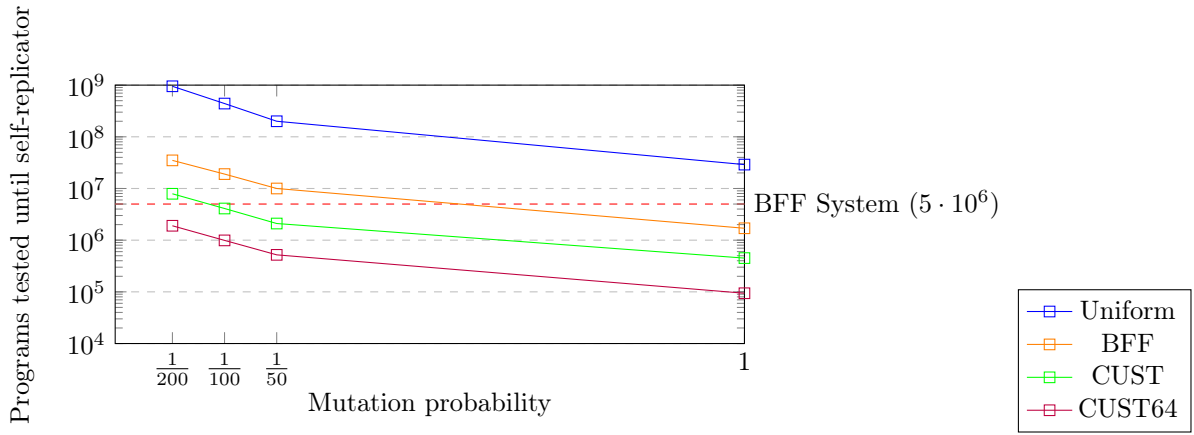
\begin{figure}[h!]
    \centering
\begin{tikzpicture}
\begin{axis}[
    width=0.6\textwidth,
    height=5cm,
    xlabel={Mutation probability},
    ylabel={Programs tested until self-replicator},
    xmin=0, xmax=1,
    ymin=10000, ymax=1000000000,
    xtick={0.0005,0.001,0.002,1},
    ytick={10000,100000,1000000,10000000,100000000,1000000000},
    xticklabels={$\frac{1}{200}$,$\frac{1}{100}$,$\frac{1}{50}$,1},
    extra y ticks={5000000},
    extra y tick style={grid=major,major grid style=red,ticklabel pos=right},
    extra y tick labels={BFF System ($5\cdot10^6$)},
    legend to name=named,
    ymajorgrids=true,
    ymode=log,
    xmode=log,
    grid style=dashed,
]

\addplot[
    color=blue,
    mark=square,
    ]
    coordinates {
    (0.0005,950000000)(0.001,440000000)(0.002,200000000)(1.0,29000000)
    };
\addplot[
    color=orange,
    mark=square,
    ]
    coordinates {
    (0.0005,35000000)(0.001,19000000)(0.002,10000000)(1.0,1700000)
    };
\addplot[
    color=green,
    mark=square,
    ]
    coordinates {
    (0.0005,7900000)(0.001,4100000)(0.002,2100000)(1.0,450000)
    };
\addplot[
    color=purple,
    mark=square,
    ]
    coordinates {
    (0.0005,1900000)(0.001,990000)(0.002,520000)(1.0,94000)
    };
    \legend{Uniform,BFF,CUST,CUST64}
    
\end{axis}

\end{tikzpicture}
    \ref{named}
    \caption{Programs tested until we find the first self-replicator}
    \label{fig:plot_muts}
\end{figure}

In Figure~\ref{fig:plot_muts} we plot the durations from Table~\ref{tab:replicator_time} for each of the distributions. One can clearly see they behave in a similar pattern, just shifted. With increasing mutation rate the time until we find a self-replicator goes down for all the distributions. If we constrain ourselves to mutation probabilities that are on the lower range of what we see in the BFF system, we can still outperform the BFF system if we choose the right distribution (using $\mathcal{CUST}64$ beats running the BFF system even for $p=1/200$).


From these experiments, we can conclude that the BFF system is not better than our simple random process at finding self-replicators. It is left to ask whether running the BFF system exhibits other interesting properties.

\subsection{Blocking experiments}
The idea of having more and more complex programs appear over time echoes biological evolution where more and more complex life forms evolve from previous ones.
A natural way to test for compositionality in a system is to limit the depth of a \textit{merger}, defined as a consecutive copy of at least two bytes that were not copied together before. These mergers, over the history of the soup, form a tree. 

The \emph{depth} of a merger is the depth it has in the merger tree. Limiting this depth means that we forbid the BFF system from completing mergers above a specific depth: if such a merger happens, we cancel the interaction and return the programs to the soup in their original state. Similarly, we can define the \emph{width} of a merger to be the number of unique parents a string\footnote{The tree looks at strings that are created with copy operations of consecutive source bytes instead of complete programs.} has in this tree, and we can ask what happens if we prohibit mergers above a certain width.
Both notions capture the complexity of the strings arising from the BFF process. 

\begin{figure}[t!]
\begin{minipage}[t]{0.5\linewidth}

    \centering
    \includegraphics[width=\linewidth]{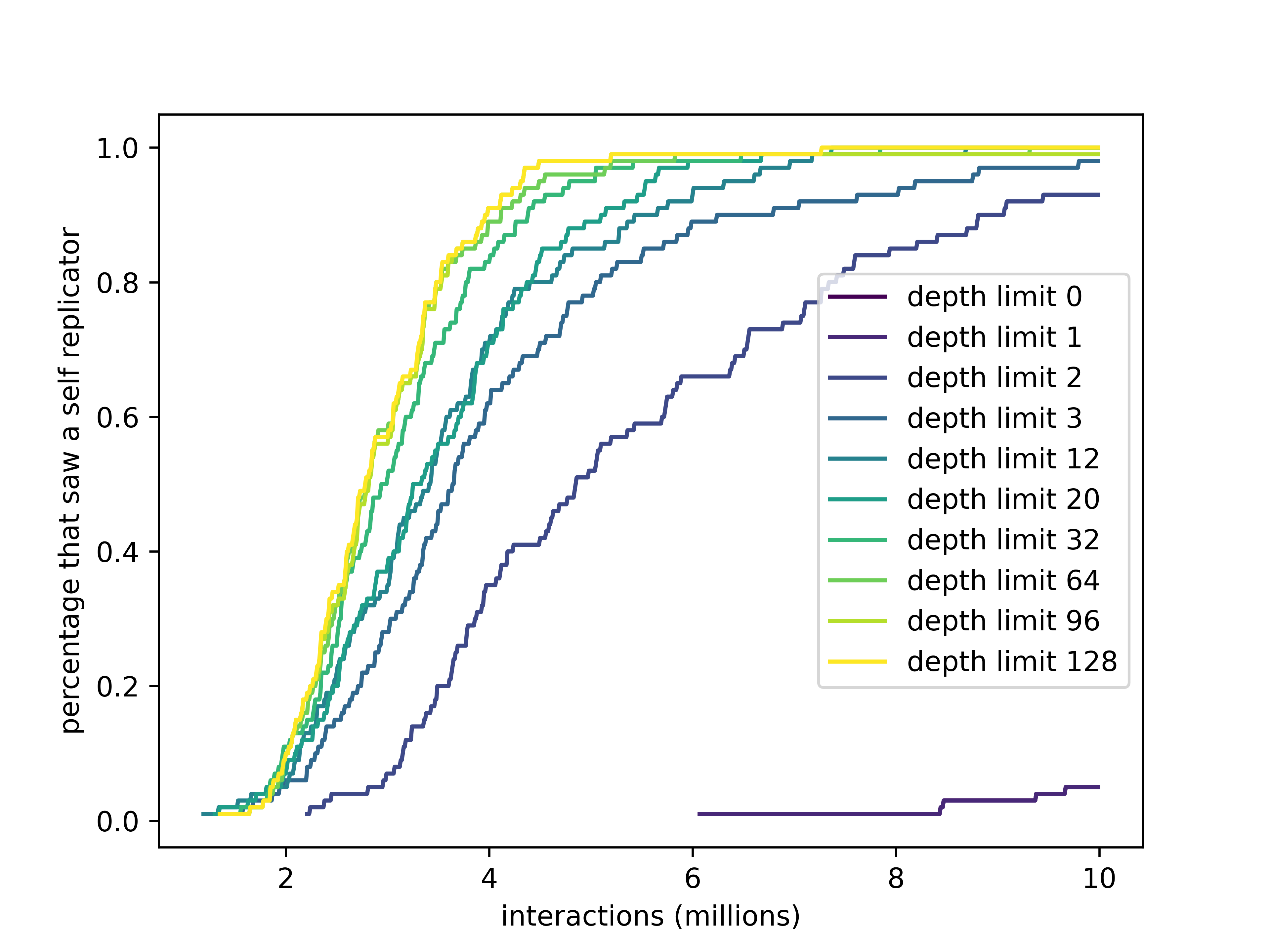}
\end{minipage}%
\begin{minipage}[t]{0.5\linewidth}
    \centering

        \includegraphics[width=\linewidth]{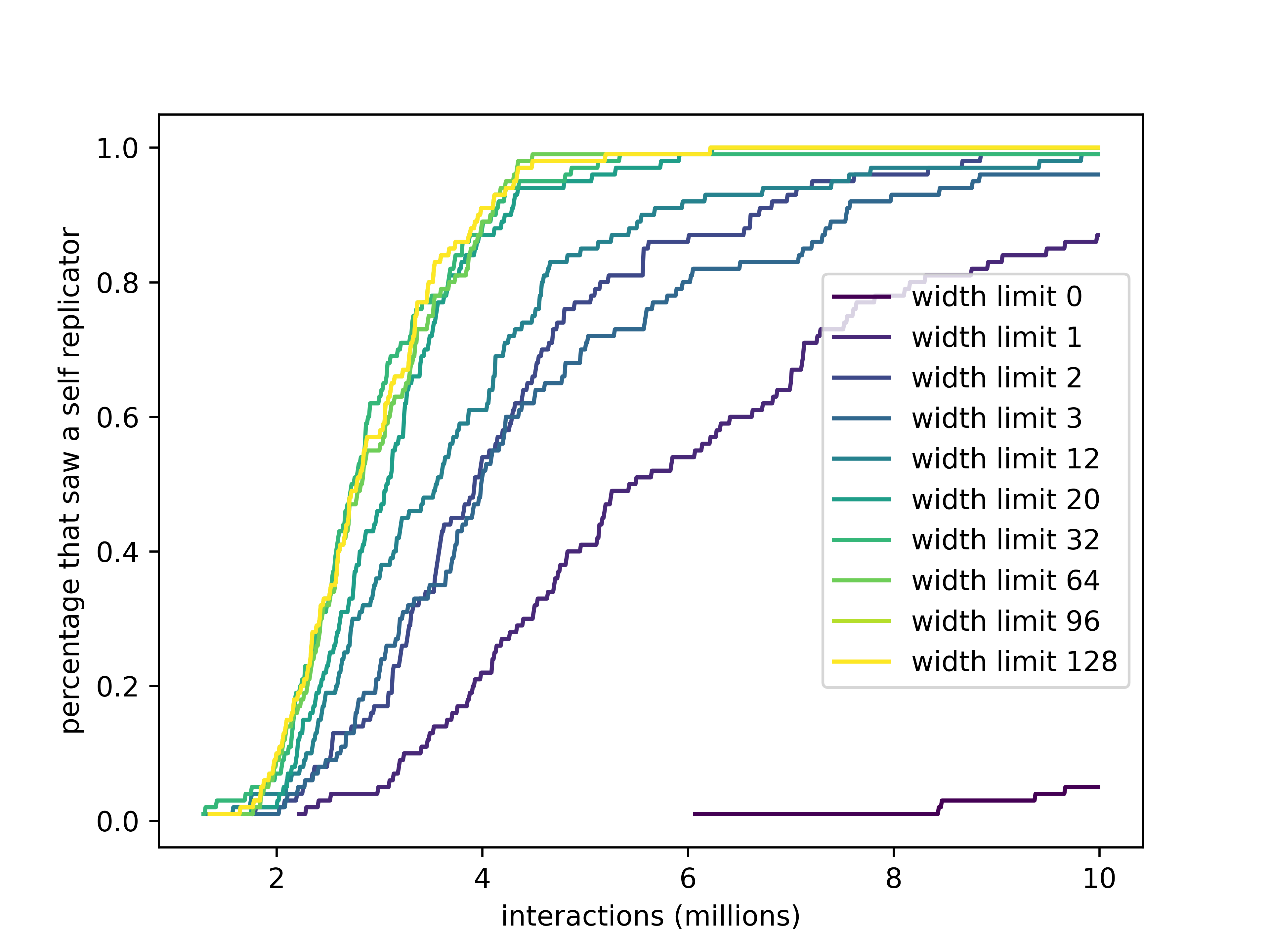}
\end{minipage}
    \caption{Percentages of runs that see a self-replicator with different restrictions for mergers}
        \label{fig:depthwidth}

\end{figure}

The idea of a compositional structure, and with it the essentialness of merger depth or width for the discovery of self-replicators, in a sense contradicts the results we saw in Sections~\ref{sec:sampling} and \ref{sec:bytes}. Randomness is not able to construct mergers, as every program develops independently. 
To give a conclusive answer to the question of whether randomness explains everything that is going on in the BFF system, we analyze the setup with limited merger depth or width in more detail.

In Figure~\ref{fig:depthwidth}, we track the time until the first self-replicator appears, with various limits on depth and width. We see that emergence of self-replicators is quite robust to merger blocking, with the process merely slows down somewhat when the depth or width limit is as low as 2. 

Note that limit $0$ (and also $1$ in case of depth) is special: it forbids any kind of copy, which means that the only way to find self-replicators is to use {\tt+} and {\tt-} to modify the bytes. This is a restriction that goes beyond just blocking mergers: in particular, it is significantly more restrictive than the random walk setting. This system still finds self-replicators, it is just significantly slower. When running for $50$ million epochs, the percentage of programs that find a self-replicator already increases to $40\%$.

This gives evidence for the hypothesis that limiting the width or depth of a merger does not actually prevent the emergence of self-replicators, but rather prevents them from taking over the soup. Thus compositionality is not needed for the discovery of self-replicators in the BFF system.

\section*{Acknowledgments}

We thank Vassilis Papadopoulos for bringing the issue of randomness being fast at solving self-replicators to our attention and for the fruitful discussions.

\bibliographystyle{plain} 
\bibliography{sources}

@article{alakuijala2024computational,
  title={Computational life: How well-formed, self-replicating programs emerge from simple interaction},
  author={Agüera y Arcas, Blaise and Alakuijala, Jyrki and Evans, James and Laurie, Ben and Mordvintsev, Alexander and Niklasson, Eyvind and Randazzo, Ettore and Versari, Luca},
  journal={arXiv preprint arXiv:2406.19108},
  year={2024}
}

@misc{bf93,
    key = {Müller, Urban},
    title = {Müller, {U}rban. \url{dev/lang/brainfuck-2.lha, 1993. }{https://aminet.net/package/dev/lang/brainfuck-2}},
    note = {[Online; accessed 24-Feb-2026]}
}

@misc{cubff,
    key = {Agüera y Arcas, Blaise and Alakuijala, Jyrki and Evans, James and Laurie, Ben and Mordvintsev, Alexander and Niklasson, Eyvind and Randazzo, Ettore and Versari, Luca},
    title = {Self-replication detector in the {CUBFF} Codebase \url{}{https://github.com/paradigms-of-intelligence/cubff/blob/main/common{\_}language.h}},
    note = {[Online; accessed 24-Feb-2026]}
}

@misc{bffselfmove,
    key = {Agüera y Arcas, Blaise and Alakuijala, Jyrki and Evans, James and Laurie, Ben and Mordvintsev, Alexander and Niklasson, Eyvind and Randazzo, Ettore and Versari, Luca},
    title = {\texttt{bff{\_}selfmove} in the {CUBFF} Codebase \url{}{https://github.com/paradigms-of-intelligence/cubff/blob/main/bff{\_}selfmove.cu}},
    note = {[Online; accessed 24-Feb-2026]}
}

@article{10.1098/rsta.2016.0350,
    author = {C G, Nitash and LaBar, Thomas and Hintze, Arend and Adami, Christoph},
    title = {Origin of life in a digital microcosm},
    journal = {Philosophical Transactions of the Royal Society A: Mathematical, Physical and Engineering Sciences},
    volume = {375},
    number = {2109},
    pages = {20160350},
    year = {2017},
    month = {11},
    issn = {1364-503X},
    doi = {10.1098/rsta.2016.0350},
    url = {https://doi.org/10.1098/rsta.2016.0350},
    eprint = {https://royalsocietypublishing.org/rsta/article-pdf/doi/10.1098/rsta.2016.0350/389495/rsta.2016.0350.pdf},
}

\end{document}